\documentclass[runningheads]{llncs}
\usepackage[T1]{fontenc}
\usepackage{graphicx}

\usepackage{url}
\usepackage{algorithm}
\usepackage{algorithmic}

\usepackage{tikz}
\usetikzlibrary{arrows,shapes}
\usetikzlibrary{positioning}
\usetikzlibrary{shadows}
\usepackage{listings}
\usepackage{bm}
\usepackage{paralist}

\newcommand{\lw}[1]{\smash{\lower-8.ex\hbox{#1}}}

\newcommand{\recongo}{\textit{recongo}}
\newcommand{\pariss}{\textit{PARIS single}}
\newcommand{\aiger}{\textit{ReconfAIGERation}}
\newcommand{\parisp}{\textit{PARIS}}

\newcommand{\clingo}{\textit{clingo}}
\newcommand{\code}[1]{\lstinline[basicstyle=\ttfamily]{#1}}

\begin{document}
\title{Bounded Combinatorial Reconfiguration with Answer Set Programming}
\titlerunning{recongo: Bounded Combinatorial Reconfiguration with ASP}
\author{Yuya Yamada\inst{1} \and
  Mutsunori Banbara\inst{1}\orcidID{0000-0002-5388-727X}\and
  Katsumi Inoue\inst{2}\orcidID{0000-0002-2717-9122} \and
  Torsten Schaub\inst{3}\orcidID{0000-0002-7456-041X}}
\authorrunning{Y.Yamada et al.}
\institute{Nagoya University,
  Furo-cho, Chikusa-ku, Nagoya, 464-8601, Japan\\  
  \email{\{yuya.yamada,banbara\}@nagoya-u.jp}\and
  National Institute of Informatics,
  Hitotsubashi, Chiyoda-ku, Tokyo 101-8430, Japan\\  
  \email{inoue@nii.ac.jp} \and
  Universit{\"a}t Potsdam,
  August-Bebel-Strasse, D-14482 Potsdam, Germany\\  
  \email{torsten@cs.uni-potsdam.de}}
\maketitle              \begin{abstract}
We develop an approach called
\emph{bounded combinatorial reconfiguration}
for solving combinatorial reconfiguration
problems based on Answer Set Programming (ASP).
The general task is to study the solution spaces of source
combinatorial problems and
to decide whether or not there are sequences of feasible
solutions that have special properties.
The resulting {\recongo} solver 
covers all metrics of the solver track in the most recent
international competition on combinatorial reconfiguration
(CoRe Challenge 2022).
{\recongo} ranked first in the shortest metric of the single-engine solvers track.
In this paper, we present the design and implementation of 
bounded combinatorial reconfiguration, and
present an ASP encoding of the independent set reconfiguration problem
that is one of the most studied combinatorial reconfiguration problems.
Finally, we present empirical analysis considering all instances of 
CoRe Challenge 2022.

\end{abstract}

\section{Introduction}\label{sec:intro}

\emph{Combinatorial reconfiguration}~\cite{core:Heuvel13,core:ItoDHPSUU11,core:Nishimura18}
aims at analyzing the structure and properties
(e.g., connectivity and reachability)
of the solution spaces of source combinatorial problems.
Each solution space has a graph structure in which each node represents
an individual feasible solution, and the edges are defined by a
certain adjacency relation.
\emph{Combinatorial Reconfiguration Problems} (CRPs)
are defined in general as the task of deciding, 
for a given source problem and two among its feasible solutions,
whether or not one is reachable from another via a 
sequence of adjacent feasible solutions.
A CRP is \emph{reachable} if there exists such a sequence,
otherwise \emph{unreachable}.
\emph{CRP solvers} are computer programs which solve
combinatorial reconfiguration problems.
The solvers output a reconfiguration sequence as a solution if
reachable.

A great effort has been made to investigate the theoretical
aspects of CRPs in the field of theoretical computer science over the
last decade. For many NP-complete source problems,
their reconfigurations have been shown to be PSPACE-complete,
including
SAT reconfiguration~\cite{core:sat:GopalanKMP09,monipara17},
independent set reconfiguration~\cite{core:ItoDHPSUU11,itkaonsuueya14,kamemi12},
dominating set reconfiguration~\cite{HaddadanIMNOST16,SuzukiMN16},
graph coloring reconfiguration~\cite{core:gcp:BonsmaC09,core:gcp:CerecedaHJ11,brmcmono16,core:gcp:Itkade12},
clique reconfiguration~\cite{core:clique:itonot15},
Hamiltonian cycle reconfiguration~\cite{DBLP:journals/algorithms/Takaoka18}, and
set covering reconfiguration~\cite{core:ItoDHPSUU11}.
However, little attention has been paid so far to its practical aspect.
To stimulate research and development on practical CRP solving,
the first international 
combinatorial reconfiguration competition
(CoRe Challenge 2022; \cite{soh22:core-challenge})
has been held last year.

In this paper, we describe an approach for solving combinatorial reconfiguration
problems based on Answer Set Programming (ASP; \cite{baral02a,gellif88b,niemela99a}).
The resulting {\recongo} solver reads an CRP instance and converts it
into ASP facts. 
In turn, these facts are combined with an ASP encoding 
for CRP solving,
which are afterward solved by 
efficient ASP solvers, in our case {\clingo}~\cite{PotasscoUserGuide}.
To show the effectiveness of our approach, we conduct experiments
on the benchmark set of CoRe Challenge 2022.
 
ASP is a declarative programming paradigm for
knowledge representation and reasoning.
The declarative approach of ASP has
distinct advantages.
First, ASP provides an expressive language and is well suited for
modeling combinatorial (optimization) problems in artificial
intelligence and computer science~\cite{ergele16a}.
Second, the extension to their reconfiguration problems
can be easily done.
Finally, recent advances in 
multi-shot ASP solving~\cite{gekakasc19,karoscwa23}
allow for efficient reachability checking of combinatorial
reconfiguration problems.

The main contributions of our paper are as follows:
\begin{enumerate}
\item We present an approach 
  called \emph{bounded combinatorial reconfiguration}
  for solving combinatorial reconfiguration problems.
  Our declarative approach
  is inspired by bounded model checking~\cite{biere09},
  which is widely used in formal verification of finite state transition
  systems.
\item We develop an ASP-based CRP solver {\recongo} with the help of
  {\clingo}'s multi-shot ASP solving capacities~\cite{karoscwa23}.
  {\recongo} ranked first at the shortest metric of the single-engine
  solvers track in the CoRe Challenge 2022, 
  and ranked second or third in the other four metrics.
\item We present an ASP encoding for solving 
  the independent set reconfiguration problem.
  This problem is one of the most studied combinatorial reconfiguration
  problems that has been shown to be PSPACE-complete~\cite{core:ItoDHPSUU11}.
\item Our empirical analysis considers all 369 instances
  publicly available from the CoRe Challenge 2022 website.\footnote{\url{https://core-challenge.github.io/2022/}}
  We address the competitiveness of 
  our declarative approach by contrasting it
  to other approaches.
\end{enumerate}
Overall, the proposed declarative approach can make a
significant contribution to the state-of-the-art of CRP solving
as well as ASP application to combinatorial reconfiguration.

In the sequel, we assume some familiarity with ASP's basic
language constructs and its extension to multi-shot ASP solving.
A comprehensive explanation of 
ASP solving can be found in \cite{gekakasc12a}.
Our encodings are given in the language of {\clingo}~\cite{PotasscoUserGuide}.

 \section{Background}\label{sec:background}

\begin{figure}[t]
  \centering
  \begin{tabular}[t]{ccccccc}
    $X_{0}$ && $X_{1}$ && $X_{2}$ && $X_{3}$ \\
    \scalebox{0.6}{\begin{tikzpicture}[x=1.5cm, y=1.5cm]
\tikzset{node/.style={circle,draw=black,minimum size=1cm}}
 
\definecolor{yellow}{RGB}{255,251,0}
 
\node[node, fill=yellow!70] at (-1,1) (node1) {\textbf{1}};
  \node[node, fill=yellow!70] at (1,1) (node2) {\textbf{2}};
  \node[node] at (-1,0) (node3) {\textbf{3}};
  \node[node, fill=yellow!70] at (0,0) (node4) {\textbf{4}};
  \node[node] at (1,0) (node5) {\textbf{5}};
  \node[node] at (-1,-1) (node6) {\textbf{6}};
  \node[node] at (0,-1) (node7) {\textbf{7}};
  \node[node] at (1,-1) (node8) {\textbf{8}};
 
  \foreach \u / \v in {node1/node3, node2/node5, node3/node4, node3/node6, node4/node5,
    node5/node8, node6/node7, node7/node8}
  \draw (\u) -- (\v);
\end{tikzpicture}
 } &
    \lw{$\Rightarrow$} &
    \scalebox{0.6}{\begin{tikzpicture}[x=1.5cm, y=1.5cm]
\tikzset{node/.style={circle,draw=black,minimum size=1cm}}
 
\definecolor{yellow}{RGB}{255,251,0}
 
\node[node, fill=yellow!70] at (-1,1) (node1) {\textbf{1}};
  \node[node] at (1,1) (node2) {\textbf{2}};
  \node[node] at (-1,0) (node3) {\textbf{3}};
  \node[node, fill=yellow!70] at (0,0) (node4) {\textbf{4}};
  \node[node] at (1,0) (node5) {\textbf{5}};
  \node[node] at (-1,-1) (node6) {\textbf{6}};
  \node[node, fill=yellow!70] at (0,-1) (node7) {\textbf{7}};
  \node[node] at (1,-1) (node8) {\textbf{8}};
 
  \foreach \u / \v in {node1/node3, node2/node5, node3/node4, node3/node6, node4/node5,
    node5/node8, node6/node7, node7/node8}
  \draw (\u) -- (\v);
\end{tikzpicture} } &
    \lw{$\Rightarrow$} &
    \scalebox{0.6}{\begin{tikzpicture}[x=1.5cm, y=1.5cm]
\tikzset{node/.style={circle,draw=black,minimum size=1cm}}
 
\definecolor{yellow}{RGB}{255,251,0}
 
\node[node, fill=yellow!70] at (-1,1) (node1) {\textbf{1}};
  \node[node] at (1,1) (node2) {\textbf{2}};
  \node[node] at (-1,0) (node3) {\textbf{3}};
  \node[node] at (0,0) (node4) {\textbf{4}};
  \node[node, fill=yellow!70] at (1,0) (node5) {\textbf{5}};
  \node[node] at (-1,-1) (node6) {\textbf{6}};
  \node[node, fill=yellow!70] at (0,-1) (node7) {\textbf{7}};
  \node[node] at (1,-1) (node8) {\textbf{8}};
 
  \foreach \u / \v in {node1/node3, node2/node5, node3/node4, node3/node6, node4/node5,
    node5/node8, node6/node7, node7/node8}
  \draw (\u) -- (\v);
\end{tikzpicture}
 } &
    \lw{$\Rightarrow$} &
    \scalebox{0.6}{\begin{tikzpicture}[x=1.5cm, y=1.5cm]
\tikzset{node/.style={circle,draw=black,minimum size=1cm}}
 
\definecolor{yellow}{RGB}{255,251,0}
 
\node[node] at (-1,1) (node1) {\textbf{1}};
  \node[node] at (1,1) (node2) {\textbf{2}};
  \node[node, fill=yellow!70] at (-1,0) (node3) {\textbf{3}};
  \node[node] at (0,0) (node4) {\textbf{4}};
  \node[node, fill=yellow!70] at (1,0) (node5) {\textbf{5}};
  \node[node] at (-1,-1) (node6) {\textbf{6}};
  \node[node, fill=yellow!70] at (0,-1) (node7) {\textbf{7}};
  \node[node] at (1,-1) (node8) {\textbf{8}};
 
  \foreach \u / \v in {node1/node3, node2/node5, node3/node4, node3/node6, node4/node5,
    node5/node8, node6/node7, node7/node8}
  \draw (\u) -- (\v);
\end{tikzpicture} } \\
    start state &&&&&& goal state \\
  \end{tabular}
  \caption{An ISRP example}
  \label{fig:ex_isrp}
\end{figure}
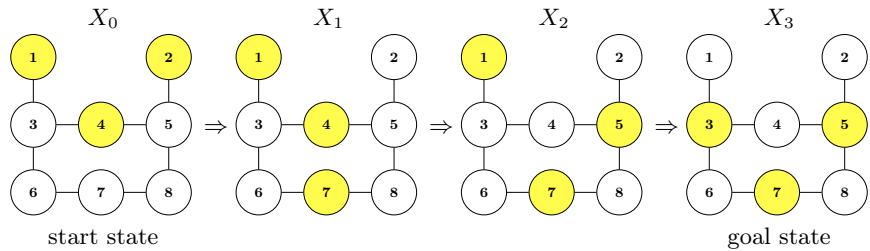

The combinatorial reconfiguration problem (CRP) is defined as
the task of deciding,
for a given source combinatorial problem
and two of its feasible solutions $X_s$ and $X_g$,
whether or not there are sequences of transitions:
\begin{equation}
  \label{eq:recocnf_seqence}
  X_{s}=X_{0}\rightarrow X_{1}\rightarrow X_{2}\rightarrow\cdots\rightarrow X_{\ell}=X_{g},
\end{equation}
where $X_s$ and $X_g$ are optional.
Each state $X_{i}$ represents a feasible solution of the source problem.
We refer to $X_s$ and $X_g$ as the start and the goal states, respectively.
We write $X\rightarrow X'$ if state $X$ at step $t$ can be followed
by state $X'$ at step $t+1$ subject to a certain \emph{adjacency relation}.
We refer to the sequence (\ref{eq:recocnf_seqence}) as
a \emph{reconfiguration sequence}.
The \emph{length} of the reconfiguration sequence, denoted by 
the symbol $\ell$, is the number of transitions
(i.e., the number of steps minus 1).
Regarding the reconfiguration sequences,
combinatorial reconfiguration problems can be 
classified into three categories:
\emph{existent}, 
\emph{shortest}, and 
\emph{longest}.
The existent-CRP is to decide whether or not there are reconfiguration sequences.
The shortest-CRP is to find the shortest reconfiguration sequences.
The longest-CRP is to find the longest reconfiguration sequences
that can not include any loop.

Let us consider the independent set reconfiguration problem (ISRP).
Its source is \emph{the independent set problem} that is to decide
whether or not there are independent sets in $G$ of size $k$,
for a given graph $G=(V,E)$ and an integer $k$.
A subset $V' \subseteq V$ is called an \emph{independent set} in
$G$ of size $k$ if 
$(u,v)\notin E$ for all $u,v\in V'$ and $|V'| = k$.
In the ISRP, each state $X$ in (\ref{eq:recocnf_seqence}) represents an independent set.
Regarding adjacency relations, 
we focus on one of the most studied relations called 
\emph{token jumping}~\cite{core:ItoDHPSUU11}.
Suppose that a \emph{token} is placed on each node in an independent set.
The token jumping meaning of 
$X\rightarrow X'$ is that
a single token ``jumps'' from one node in $X$ to any other node in $X'$.

Fig.~\ref{fig:ex_isrp} shows an example of ISRP.
The example consists of a graph having 8 nodes and 8 edges,
and the size of independent sets is $k=3$.
The independent sets (tokens) are highlighted in yellow.
We can observe that the goal state can be reachable from
the start state with length $\ell = 3$.
For instance, in the transition from $X_0$ to $X_1$,
a token jumps from node 2 in $X_{0}$ to node 7 in $X_{1}$.

 \section{The {\recongo} Approach}\label{sec:bcr}

\textbf{Basic design.}
Combinatorial reconfiguration problems can be readily expressed as
satisfiability problems.
Let $\bm{x} = \{x_1,x_2,\ldots,x_n\}$ and $C(\bm{x)}$ be the variables
and the constraints of a source combinatorial problem, respectively.
For its reconfiguration problem,
each state $X$ at step $t$ can be represented
by a set of variables
$\bm{x}^{t} = \{x_1^t,x_2^t,\ldots,x_n^t\}$.
Each adjacent relation can be represented by
a set of constraints $T(\bm{x}^{t-1},\bm{x}^{t})$ that must be satisfied.
Optionally, additional constraints
$S(\bm{x}^{0})$ and $G(\bm{x}^{\ell})$
can be added to specify conditions
on the start state $X_{s}$ and/or the goal state $X_{g}$ respectively,
as well as any other constraints that we want to enforce.
Then, the existence of a reconfiguration sequence
(\ref{eq:recocnf_seqence})
of bounded length $\ell$ is equivalent to the following 
satisfiability problem
\begin{equation}\label{eq:bcr}
\varphi_{\ell} =
 S(\bm{x}^{0})\land
 \bigwedge_{t=0}^{\ell} C(\bm{x}^{t})\land 
 \bigwedge_{t=1}^{\ell} T(\bm{x}^{t-1},\bm{x}^{t})\land
 G(\bm{x}^\ell).
\end{equation} 
We use $\varphi_{\ell}$ to check properties of
a reconfiguration relation (a transition relation) between the
possible feasible solutions of the source combinatorial problem.
We call this general framework 
``bounded combinatorial reconfiguration'',
because we consider only reconfiguration sequences
that have a bounded length $\ell$.

For reachability checking,
if $\varphi_{\ell}$ is satisfiable, 
there is a reconfiguration sequence of length $\ell$.
Otherwise, we keep on reconstructing a successor (e.g., $\varphi_{\ell+1}$)
and checking its satisfiability until 
a reconfiguration sequence is found.
Bounded combinatorial reconfiguration 
is an incomplete method, 
because it can find reconfiguration sequences if they exist,
but cannot prove unreachability in general.
However, it can be a complete method if
the diameters of solution spaces are given.
Any off-the-shelf satisfiability solvers, such as SAT solvers and CSP solvers,
can be used as back-end.
In this paper, we make use of ASP solvers, in our case {\clingo}.

\begin{algorithm}[htbp]\caption{Grounding-conscious bounded combinatorial reconfiguration}
\label{alg:improved}
\textbf{Input} $P$: problem instance of ASP fact format\\
\textbf{Input} $S(\bm{x}^{0})$, $C(\bm{x}^{t})$, $T(\bm{x}^{t-1},\bm{x}^{t})$, $G(\bm{x}^{t})$: logic program\\
\textbf{Parameter} $I^{min}$: the minimum number of steps [1]\\
\textbf{Parameter} $I^{max}$: the maximum number of steps [$none$]\\
\textbf{Parameter} $I^{stop}$: termination criterion [\textsf{SAT}]\\
\textbf{Parameter} $I^{search}$: path search [\textsf{shortest}]\\[-1em]
\begin{algorithmic}[1] \STATE $ctl\gets \textrm{create an object of ASP solver}$
\STATE $i\gets 0$
\STATE $ret\gets none$
\STATE $model\gets none$
\IF {$I^{search} = \textsf{longest}$}
\STATE $I^{min}\gets I^{max}$
\ENDIF
\STATE add a statement ``\texttt{\#external query($t$).}'' to $G(\bm{x}^{t})$
\WHILE{($I^{max} = none$ \OR $i < I^{max}$) \AND\\
       ($I^{min} = none$ \OR $i < I^{min}$ \OR $ret = none$ \OR $ret \neq I^{stop}$)}
\STATE $parts \gets \textrm{an empty list}$
\STATE $parts.append(C(\bm{x}^{i}))$
\STATE $parts.append(G(\bm{x}^{i}))$
\IF {$i > 0$}
\STATE $parts.append(T(\bm{x}^{i-1},\bm{x}^{i}))$
\STATE $ctl.release\_external(\texttt{query(}i-1\texttt{)})$ \COMMENT{deactivating $G(\bm{x}^{i-1})$}
\ELSE
\STATE $parts.append(P)$
\STATE $parts.append(S(\bm{x}^{0}))$
\ENDIF
\STATE $ctl.ground(parts)$
\STATE $ctl.assign\_external(\texttt{query(}i\texttt{)}, \TRUE)$ \COMMENT{activating $G(\bm{x}^{i})$}
\STATE $(ret,\ model) = ctl.solve()$
\STATE $i \gets i+1$
\ENDWHILE
\STATE $ctl.close()$
\IF{$model \neq none$}
\PRINT \textsf{REACHABLE}
\ELSIF{$I^{max} \neq none$ \AND $i \geq I^{max}$}
\PRINT{\textsf{UNREACHABLE}}
\ELSE
\PRINT{\textsf{UNKNOWN}}
\ENDIF
\end{algorithmic}
\end{algorithm}

\textbf{Algorithm.}
We present an algorithm of bounded combinatorial reconfiguration.
Obviously it is inefficient to fully reconstruct $\varphi_{\ell}$ in
each transition because of the expensive grounding.
Instead, we incrementally construct $\varphi_{\ell}$ from its 
predecessor $\varphi_{\ell-1}$ by adding 
$C(\bm{x}^{\ell})$, $T(\bm{x}^{\ell-1},\bm{x}^{\ell})$, and $G(\bm{x}^{\ell})$
and deactivating $G(\bm{x}^{\ell -1})$.
This can be easily done by utilizing {\clingo}'s multi-shot ASP
solving capacities~\cite{karoscwa23}.

Algorithm~\ref{alg:improved} shows the pseudo code of a grounding-conscious
algorithm for bounded combinatorial reconfiguration.
We use five variables at the loop in Lines 9--24
to control the successive grounding and solving
of an input logic program.
The input consists of a problem instance $P$ of ASP fact format
and four subprograms
$S(\bm{x}^{0})$, $C(\bm{x}^{t})$, $T(\bm{x}^{t-1},\bm{x}^{t})$,
and $G(\bm{x}^{t})$.
The values of $I^{min}$ and $I^{max}$ respectively indicate
the minimum number of steps (1 by default) and
the maximum number of steps ($none$ by default).
A variable $I^{stop}$ is used to specify a termination criterion
(\textsf{SAT} by default or \textsf{UNSAT}).
The value of $i$ indicates each step, and
a variable $ret$ is used to store the solving result.
In addition,
$I^{search}$ is used to switch a search type
(\textsf{shortest} by default, \textsf{longest}, or \textsf{existent}),
and $model$ is used to store the answer sets.
We note that the external atom \texttt{query($t$)} in Line 8 is used to
deactivate $G(\bm{x}^{t})$ as well as to activate it.

In each step of the loop, 
the subprograms stored in the list $parts$ are grounded and solved.
For instance, in step $i=0$,
the $ctl.ground(parts)$ method in Line 20 grounds 
$P\land S(\bm{x}^{0})\land C(\bm{x}^{0})\land G(\bm{x}^{0})$
in which \texttt{query(0)} is set to true in Line 21 in order to 
activate $G(\bm{x}^{0})$.
Then, the method $ctl.solve()$ in Line 22 checks its satisfiability,
that is, whether or not the goal state is reachable from the start state.
Finally, the value of step $i$ is incremented by 1.

This process iterates until the termination criterion is met. 
Each following step checks the satisfiability of the logic program
$P\land S(\bm{x}^{0})\land
\bigwedge_{t=0}^{i} C(\bm{x}^{t})\land 
\bigwedge_{t=1}^{i} T(\bm{x}^{i-1},\bm{x}^{i})\land
G(\bm{x}^{i})$, in which 
the current external atom
\texttt{query($i$)}
is set to true, but the previous \texttt{query($i-1$)} is permanently
set to false in Line 15.
The algorithm searches shortest
reconfiguration sequences in a default setting.
For searching longest ones, all we have to do is just
assigning the value of $I^{max}$ to $I^{min}$ in Line 6.

\begin{figure}[t]
  \thicklines
  \setlength{\unitlength}{1.28pt}
  \small
  \centering
  \scalebox{0.9}{﻿\begin{picture}(280,57)(4,-10)
  \put(  0, 20){\dashbox(50,24){\shortstack{CRP instance}}}
  \put( 60, 20){\framebox(50,24){converter}}
  \put(120, 20){\dashbox(50,24){\shortstack{ASP facts}}}
  \put(120,-10){\dashbox(50,24){\shortstack{logic program}}}
  \put(180,-10){\framebox(50,54){}}
  \put(185, 29){\framebox(40,12){\clingo}}
  \put(185, -7){\framebox(40,24){\shortstack{BCR\\ algorithm}}}
  \put(240, 20){\dashbox(50,24){\shortstack{CRP solution}}}
  \put( 50, 32){\vector(1,0){10}}
  \put(110, 32){\vector(1,0){10}}
  \put(170, 32){\vector(1,0){10}}
  \put(230, 32){\vector(1,0){10}}
  \put(170, +2){\line(1,0){4}}
  \put(174, +2){\line(0,1){30}}
  \put(195,  17){\vector(0,1){12}}
  \put(215, 29){\vector(0,-1){12}}
\end{picture}  

 }
  \caption{The architecture of {\recongo}}
  \label{fig:arch}
\end{figure}
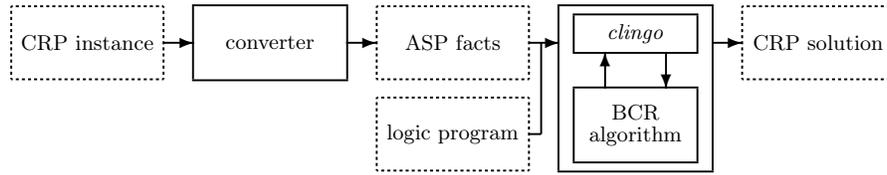

\textbf{Implementation.}
Bounded combinatorial reconfiguration (BCR)
in Algorithm~\ref{alg:improved}
can be easily implemented using the latest {\clingo}'s
Python API.~\footnote{\url{https://potassco.org/clingo/python-api/current/}}
The resulting {\recongo} solver is a general-purpose CRP solver.
The architecture of {\recongo} is shown in Fig.~\ref{fig:arch}.
{\recongo} reads an CRP instance and converts it
into ASP facts. 
In turn, these facts are combined with an ASP encoding 
for CRP solving,
which are afterward solved by
the BCR algorithm powered by {\clingo}.
{\recongo} covers all metrics of the solver track in the most recent
international competition on combinatorial reconfiguration
(CoRe Challenge 2022): 
\emph{existent}, 
\emph{shortest}, and 
\emph{longest}.

 \section{ASP encoding for Independent Set Reconfiguration}\label{sec:encoding}

\begin{lstlisting}[float=t,caption={ASP fact format of ISRP instance in Fig.~\ref{fig:ex_isrp}},captionpos=b,frame=single,label=code:isrp_example.lp,numbers=none,breaklines=true,columns=fullflexible,keepspaces=true,basicstyle=\ttfamily\scriptsize\footnotesize]
k(3).
node(1).   node(2).   node(3).   node(4).
node(5).   node(6).   node(7).   node(8).
edge(1,3). edge(2,5). edge(3,4). edge(3,6).
edge(4,5). edge(5,8). edge(6,7). edge(7,8).
start(1).  start(2).  start(4).
goal(3).   goal(5).   goal(7).
\end{lstlisting}
\begin{lstlisting}[float=t,caption={ASP encoding for ISRP solving},captionpos=b,frame=single,label=code:isrpTJ_exact1_inc.lp,numbers=left,breaklines=true,columns=fullflexible,keepspaces=true,basicstyle=\ttfamily\scriptsize\footnotesize]
#program base.
% start state
:- not in(X,0), start(X).

#program step(t).
% independent set constraints
K { in(X,t): node(X) } K :- k(K).
:- in(X,t), in(Y,t), edge(X,Y).

% adjacency relation: token jumping
moved_from(X,t) :- in(X,t-1), not in(X,t), t > 0.
:- not 1 { moved_from(X,t) } 1, t > 0.

#program check(t).
% goal state
:- not in(X,t), goal(X), query(t).
\end{lstlisting}
\begin{lstlisting}[float=t,caption={No loop constraints for ISRP solving},captionpos=b,frame=single,label=code:hint_noloop_inc.lp,numbers=none,breaklines=true,columns=fullflexible,keepspaces=true,basicstyle=\ttfamily\scriptsize\footnotesize]
#program step(t).
:- in(X,t) : in(X,T); not in(X,t) : not in(X,T),node(X); T = 0..t-1; t>0.
\end{lstlisting}

We present an ASP encoding for solving the independent set
reconfiguration problem (ISRP).

\textbf{Fact format.}
The input of ISRP is an independent set problem,
a start state, and a goal state.
Listing~\ref{code:isrp_example.lp} shows an ASP fact format of the ISRP
instance in Fig.~\ref{fig:ex_isrp}.
The predicate \code{k/1} represents the size of independent sets.
The predicates \code{node/1} and \code{edge/2}
represent the nodes and edges, respectively.
The predicates \code{start/1} and \code{goal/1} represent
the independent sets of the start and goal states, respectively.
For instance, the atom \code{start(4)} means that
the node \code{4} is in an independent set at the start state.

\textbf{First order encoding.}
Listing~\ref{code:isrpTJ_exact1_inc.lp} shows
an ASP encoding for ISRP solving.
The encoding consists of three parts:
\code{base}, \code{step(t)}, and \code{check(t)}.
The parameter \code{t} represents each step in a reconfiguration sequence.
The atom \code{in(X,t)} is intended to represent that the node
\code{X} is in an independent set at step \code{t}.
The \code{base} part specifies the constraints 
on the start state $S(\bm{x}^{0})$.
The rule in Line 3 enforces that
\code{in(X,0)} holds for each node \code{X} in the start state.
The \code{step(t)} part specifies
the constraints that must be satisfied at each step \code{t}.
The rules in Lines 7--8 represent the constraints of independent set
$C(\bm{x}^{\texttt{t}})$.
The rule in Line 7 generates a candidate independent set with size \code{K}.
The rule in Line 8 enforces that 
no two nodes connected by an edge are in an independent set.
The rules in Lines 11--12 represent the adjacency relation
$T(\bm{x}^{\texttt{t-1}},\bm{x}^{\texttt{t}})$.
The auxiliary atom \code{moved_from(X,t)} in Line 11 represents that
a token jumps from node \code{X} to any other node, from step \code{t-1} to \code{t}.
The rule in Line 12 enforces that
exactly one token jumps at each step \code{t}.
The \code{check(t)} part specifies the termination condition
that must be satisfied at the goal state $G(\bm{x}^{t})$.
The rule in Line 16 enforces that
\code{in(X,t)} holds for each node \code{X} in the goal state.
The volatility of this rule is handled by 
a truth assignment to the external atom \code{query(t)},
as explained in Algorithm~\ref{alg:improved}.
In addition, Listing~\ref{code:hint_noloop_inc.lp} shows a simple
encoding that ensure there is no loop in reconfiguration sequences.
This constraint is essential for longest-ISRP solving.

\begin{lstlisting}[float=t,caption={Hint constraints for ISRP solving},captionpos=b,frame=single,label=code:hint_constraints.lp,numbers=left,breaklines=true,columns=fullflexible,keepspaces=true,basicstyle=\ttfamily\scriptsize\footnotesize]
#program step(t).
% distance constraint (d1)
:- not { not in(X,t): start(X) } t.

% distance constraint (d2)
:- not { not in(X,T): goal(X) } t-T, T = 0..t-1, query(t).

% token constraint (t1)
:- moved_from(X,t-1), in(X,t), t > 1.
					    
% token constraint (t2)
moved_to(X,t) :- not in(X,t-1), in(X,t), t > 0.
:- moved_to(X,t-1), not in(X,t), t > 1.

% maximal independent set heuristic (h)
#heuristic in(Y,t): edge(X,Y), not in(X,t). [level_max-t,true]
#heuristic in(Y,t): edge(Y,X), not in(X,t). [level_max-t,true]
\end{lstlisting}
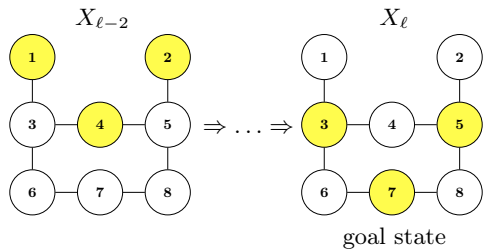
\begin{figure}[t]
  \centering
  \begin{tabular}[t]{ccc}
    $X_{\ell-2}$ && $X_{\ell}$ \\
    \scalebox{0.6}{\begin{tikzpicture}[x=1.5cm, y=1.5cm]
\tikzset{node/.style={circle,draw=black,minimum size=1cm}}
 
\definecolor{yellow}{RGB}{255,251,0}
 
\node[node, fill=yellow!70] at (-1,1) (node1) {\textbf{1}};
  \node[node, fill=yellow!70] at (1,1) (node2) {\textbf{2}};
  \node[node] at (-1,0) (node3) {\textbf{3}};
  \node[node, fill=yellow!70] at (0,0) (node4) {\textbf{4}};
  \node[node] at (1,0) (node5) {\textbf{5}};
  \node[node] at (-1,-1) (node6) {\textbf{6}};
  \node[node] at (0,-1) (node7) {\textbf{7}};
  \node[node] at (1,-1) (node8) {\textbf{8}};
 
  \foreach \u / \v in {node1/node3, node2/node5, node3/node4, node3/node6, node4/node5,
    node5/node8, node6/node7, node7/node8}
  \draw (\u) -- (\v);
\end{tikzpicture}
 } &
    \lw{$\Rightarrow \dots \Rightarrow$} &
    \scalebox{0.6}{\begin{tikzpicture}[x=1.5cm, y=1.5cm]
\tikzset{node/.style={circle,draw=black,minimum size=1cm}}
 
\definecolor{yellow}{RGB}{255,251,0}
 
\node[node] at (-1,1) (node1) {\textbf{1}};
  \node[node] at (1,1) (node2) {\textbf{2}};
  \node[node, fill=yellow!70] at (-1,0) (node3) {\textbf{3}};
  \node[node] at (0,0) (node4) {\textbf{4}};
  \node[node, fill=yellow!70] at (1,0) (node5) {\textbf{5}};
  \node[node] at (-1,-1) (node6) {\textbf{6}};
  \node[node, fill=yellow!70] at (0,-1) (node7) {\textbf{7}};
  \node[node] at (1,-1) (node8) {\textbf{8}};
 
  \foreach \u / \v in {node1/node3, node2/node5, node3/node4, node3/node6, node4/node5,
    node5/node8, node6/node7, node7/node8}
  \draw (\u) -- (\v);
\end{tikzpicture} } \\
    && goal state \\
  \end{tabular}
  \caption{An invalid reconfiguration sequence forbidden by \code{d2}}
  \label{fig:d2}
\end{figure}
\begin{figure}[t]
  \newcommand{\rz}[1]{\smash{\raise5.ex\hbox{#1}}}
  \centering
  \begin{tabular}[t]{ccccc}
    $X_{\ell-2}$ && $X_{\ell-1}$ && $X_{\ell}$ \\
    \scalebox{0.6}{\begin{tikzpicture}[x=1.5cm, y=1.5cm]
\tikzset{node/.style={circle,draw=black,minimum size=1cm}}
  \tikzset{wc/.style={circle,draw=white,minimum size=0.4cm}}
 
\definecolor{yellow}{RGB}{255,251,0}

\node[node, fill=yellow!70] at (-1,1) (node1) {\textbf{1}};
  \node[wc] at (-0.5,0.5) (wc1) {};
  \node[node] at (0,0) (node2) {\textbf{2}};
  \node[wc] at (0.5,0.5) (wc2) {};
  \node[node, fill=yellow!70] at (1,1) (node3) {\textbf{3}};
 
  \foreach \u / \v in {node1/wc1, wc1/node2, node2/wc2, wc2/node3}
  \draw[dashed] (\u) -- (\v);
\end{tikzpicture} } &
    \rz{$\Rightarrow$} &
    \scalebox{0.6}{\begin{tikzpicture}[x=1.5cm, y=1.5cm]
\tikzset{node/.style={circle,draw=black,minimum size=1cm}}
  \tikzset{wc/.style={circle,draw=white,minimum size=0.4cm}}
 
\definecolor{yellow}{RGB}{255,251,0}

\node[node] at (-1,1) (node1) {\textbf{1}};
  \node[wc] at (-0.5,0.5) (wc1) {};
  \node[node, fill=yellow!70] at (0,0) (node2) {\textbf{2}};
  \node[wc] at (0.5,0.5) (wc2) {};
  \node[node, fill=yellow!70] at (1,1) (node3) {\textbf{3}};
 
  \foreach \u / \v in {node1/wc1, wc1/node2, node2/wc2, wc2/node3}
  \draw[dashed] (\u) -- (\v);
\end{tikzpicture} } &
    \rz{$\Rightarrow$} &
    \scalebox{0.6}{\begin{tikzpicture}[x=1.5cm, y=1.5cm]
\tikzset{node/.style={circle,draw=black,minimum size=1cm}}
  \tikzset{wc/.style={circle,draw=white,minimum size=0.4cm}}
 
\definecolor{yellow}{RGB}{255,251,0}

\node[node, fill=yellow!70] at (-1,1) (node1) {\textbf{1}};
  \node[wc] at (-0.5,0.5) (wc1) {};
  \node[node, fill=yellow!70] at (0,0) (node2) {\textbf{2}};
  \node[wc] at (0.5,0.5) (wc2) {};
  \node[node] at (1,1) (node3) {\textbf{3}};
 
  \foreach \u / \v in {node1/wc1, wc1/node2, node2/wc2, wc2/node3}
  \draw[dashed] (\u) -- (\v);
\end{tikzpicture} }
\end{tabular}
  \caption{An invalid reconfiguration sequence forbidden by \code{t1}}
  \label{fig:t1}
\end{figure}
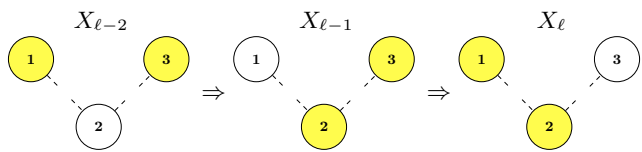

\textbf{Hint constraints.}
We present a search heuristics and four hint constraints
to accelerate ISRP solving.
Their ASP encodings are shown in Listing~\ref{code:hint_constraints.lp}.

\begin{itemize}
\item The hints \code{d1} and \code{d2} are
  constraints on the lower bound of
  distance between two states in each reconfiguration sequence.
  The hint \code{d1} in Line 3 enforces that,
  for each step \code{t},
  there are at most \code{t} nodes
  that are in the start state but not in step \code{t}.
Similarly, the hint \code{d2} in Line 6 enforces that,
  for each step \code{t} and \code{T}$\in\{$\code{0}$\ldots$\code{t-1}$\}$,
  there are at most \code{t-T} nodes
  that are in the goal state but not in step \code{T}.
For instance, Fig.~\ref{fig:d2} shows an example of
  invalid reconfiguration sequences forbidden by \code{d2}.
  In the sequence, the lower bound of distance between
  $X_{\ell-2}$ and $X_{\ell}$ is 3, but it is greater than the
  possible number of transitions, namely 2.
\item The hints \code{t1} and \code{t2} are 
  constraints that forbid redundant token moves.
  The hint \code{t1} in Line 9 enforces that,
  in two consecutive transitions,
  no token jumps back to a node \code{X} at step \code{t} from which a
  token jumped before.
Similarly, the hint \code{t2} in Lines 12--13 enforces that
  no token jumps from a node \code{X} at step \code{t} to which a
  token jumped before.
For instance, Fig.~\ref{fig:t1} shows an example of 
  invalid reconfiguration sequences forbidden by \code{t1}.

\item The heuristics \code{h} is a domain-specific heuristics that
  attempts to make each state to be a maximal independent set.
  This can be easily done by using {\clingo}'s
  \code{#heuristic} statements~\cite{gekaotroscwa13a}.
  They allow for modifying the heuristic of {\clingo} from within a
  logic program.
  In {\clingo}'s heuristic programming,
  each atom has a level, and its default value is 0.
  For each step \code{t} and for each \code{edge(X,Y)},
  the statement in Line 16 gives a higher level to the atom \code{in(Y,t)}
  if its adjacent node \code{X} is not in an independent set at step \code{t}.
  The statement in Line 17 works in the same way.
\end{itemize}
The distance constraints \code{d1} and \code{d2} are
domain-independent and can be applied to many CRPs.
In contrast, the others are domain-specific constraints for ISRP.
We note that the token hints \code{t1} and \code{t2} can
not be used for longest-ISRP solving.

 \section{Experiments}\label{sec:experiment}

\begin{table}[t]
  \tabcolsep = 2mm
  \renewcommand{\arraystretch}{1.2}
  \centering
  \caption{The number of solved instances for existent-ISRP}
  \begin{tabular}{l|r|rrrrr|r}
  & \code{all-hints} & \multicolumn{5}{c|}{\code{all-hints}} & \code{no-hint} \\
&  & \code{/t2} & \code{/t1} & \code{/h} & \code{/d2} & \code{/d1} & \\ \hline
  \textsf{REACHABLE} & \textbf{234} & 233 & 233 & 229 & 195 & 232 & 188 \\
  \textsf{UNREACHABLE} & \textbf{6} & \textbf{6} & 4 & 4 & \textbf{6} & \textbf{6} & 4 \\ \hline
  Total & \textbf{240} & 239 & 237 & 233 & 201 & 238 & 192 \\
\end{tabular}
   \label{tbl:existent_self_detail}
\end{table}
\begin{figure}[t]
  \centering
  \includegraphics[scale=0.23]{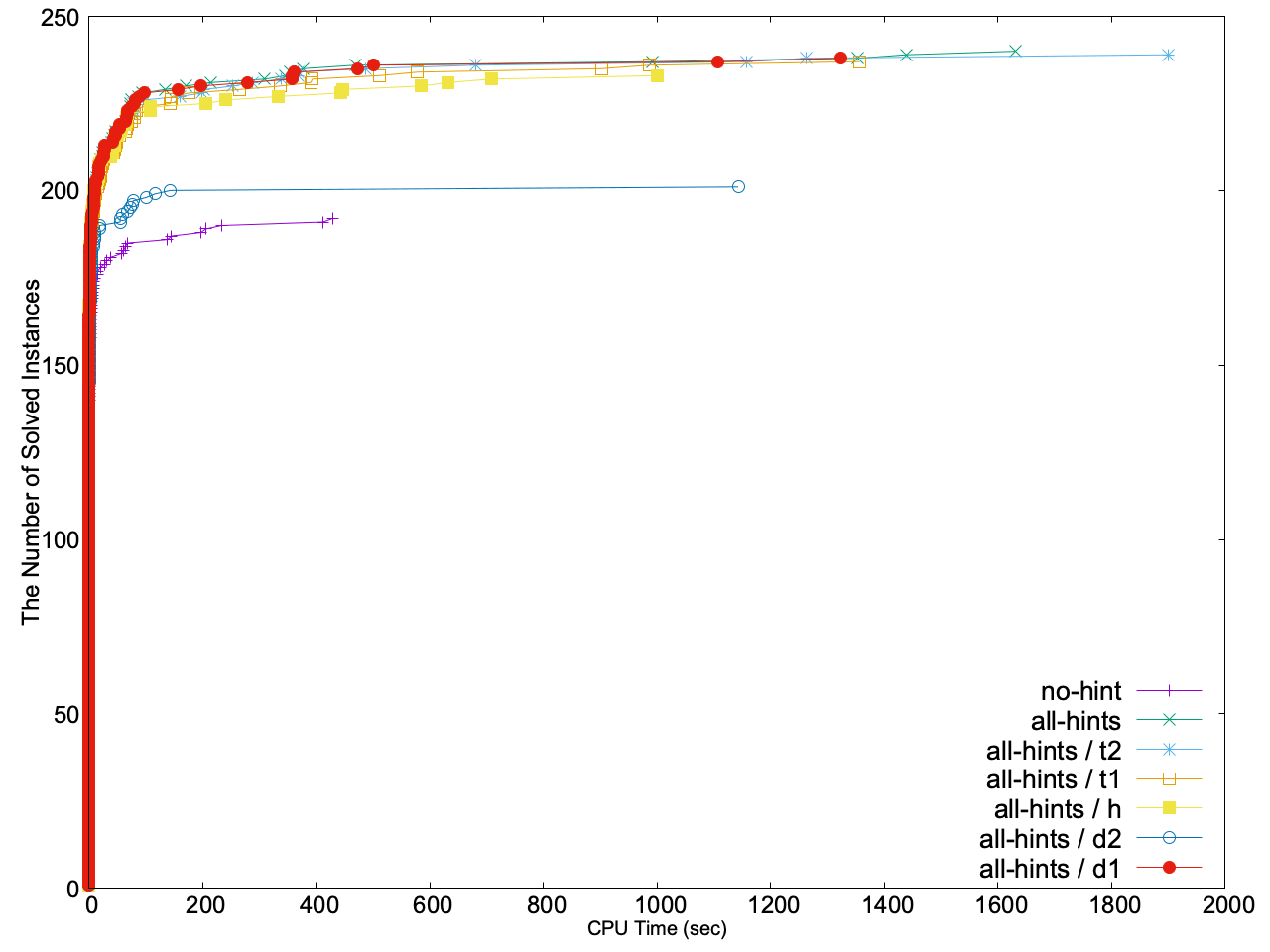}
  \caption{Cactus plot of existent-ISRP}
  \label{fig:cactus:existent}
\end{figure}
\begin{table}[t]
  \tabcolsep = 1.3mm
  \renewcommand{\arraystretch}{1.2}
  \centering
  \caption{The number of solved instances and the score for longest-ISRP}
  \begin{tabular}{l|rrrrrrrr}
  & \code{no-hint} & \code{d1d2h} & \code{d1d2} & \code{d1h}
  & \code{d2h} & \code{d1} & \code{d2} & \code{h} \\ \hline
  \#steps gained & 4,995 & \textbf{21,693} & 8,058 & 12,456 & 20,321 & 6,656 & 7,318 & 12,044\\
  \#solved instances & 189 & 221 & \textbf{224} & 194 & 213 & 213 & 223 & 191 \\\hline
\end{tabular}
   \label{tbl:longest:self}
\end{table}

To evaluate the
{\recongo} approach in Section~\ref{sec:bcr} and
the {\recongo} encoding in Section~\ref{sec:encoding},
we conduct experiments on the benchmark set of the most recent
international competition on combinatorial reconfiguration (CoRe
Challenge 2022).

Our empirical analysis considers all ISRP instances,
namely 369 in a total.
They are publicly available from CoRe Challenge 2022 website.\footnote{\url{https://core-challenge.github.io/2022/}}
This ISRP benchmark set is classified into seven families.
The benchmark family \code{color04} consists of 202 instances,
\code{grid} of 49 instances,
\code{handcraft} of 6 instances,
\code{power} of 17 instances,
\code{queen} of 48 instances,
\code{sp} of 30 instances, and
\code{square} of 17 instances.
The number of nodes ranges from 7 to 40,000.
We use the proposed encoding in Listing~\ref{code:isrpTJ_exact1_inc.lp} and
the hint constraints in Listing~\ref{code:hint_constraints.lp}.
The no loop constraints in Listing~\ref{code:hint_noloop_inc.lp}
are used only for longest-ISRP solving.
We used {\recongo} 0.2 powered by {\clingo} 5.
We note that
{\recongo} 0.2 searches shortest reconfiguration sequences when
$I^{search}$ is \textsf{existent} or \textsf{shortest}.
In our preliminary experiments, 
since the total number of all feasible solutions can be easily
obtained by {\clingo} for 23 instances, we set it to $I^{max}$ for each.
We ran our experiments on a Mac OS machine equipped with
Intel Xeon W 12-core 3.3 GHz processors and 96 GB RAM.
We imposed a time limit of 1 hours for each instance.

First, Table~\ref{tbl:existent_self_detail} shows the number of solved
instances for existent-ISRP.
The columns display in order
reachability and the number of solved instances for each encoding.
The best results are highlighted in bold.
The \code{no-hint} means the {\recongo} encoding without any hints.
The \code{all-hints} means the {\recongo} encoding with all hints.
The \texttt{all-hints/$X$} means the \code{all-hints} except $X$ where 
$X\in\{\texttt{d1},\texttt{d2},\texttt{t1},\texttt{t2},\texttt{h}\}$.
The \code{all-hints} solved the most, namely 240 instances out of 369.
The hints are highly effective
since the \code{all-hints} was able to solve 48 instances more than the \code{no-hint}.
Regarding single hint,
the distance hint \code{d2} is the most effective
since the difference of \texttt{all-hints/\texttt{d2}} from the
\code{all-hints} is the largest, namely 240-201=39.
It is followed by 240-233=7 of the heuristics \code{h} based on
the idea of the maximal independent set. 
Fig.~\ref{fig:cactus:existent} shows a cactus plot where 
the horizontal axis (x-axis) indicates CPU times in seconds, and 
the vertical axis (y-axis) indicates the number of solved instances.
The cactus plot clearly shows 
the contrast discussed above between
the \code{all-hints}, 
\texttt{all-hints/\texttt{d2}}, and 
\code{no-hint} encodings.

Second, we examine the longest-ISRP that is to find longest
reconfiguration sequences without any loop.
We here consider two evaluation criteria since finding optimal
sequences is quite an hard task.
One criterion is the sum of steps gained from the shortest length.
For instance, the instance \code{hc-square-02_01} has a shortest
sequence of length 30 and a longest sequence of length 74.
Thus, the number of steps gained is $74-30=44$.
Another criterion is the number of instances for which a least one
sequence was found.
Table~\ref{tbl:longest:self} shows the results of each encoding.
The best results are highlighted in bold.
The first row displays all possible combinations of three hints
\code{d1}, \code{d2}, and \code{h}.
It is noted that the token hints \code{t1} and \code{t2} can not be
applied to longest-ISRP.
For the former criterion, 
the {\recongo} encoding with the hints \code{d1d2h} has gain the most,
namely 21,693 steps in a total.
On the other hand, for the latter criterion, 
the {\recongo} encoding with the hints \code{d1d2} solved the most,
but less steps gained.

\begin{table}[t]
  \centering
  \tabcolsep = 1.4mm
  \renewcommand{\arraystretch}{1.1}
  \caption{The result of the single-engines solver track in CoRe Challenge 2022}
  \begin{tabular}{ll|rrr}
  metric & & 1st & 2nd & 3rd \\ \hline
  & solver name & {\pariss} & \textbf{\recongo} & @toda5603 \\
  existent & method & planning & ASP & greedy search/BMC \\
  & score & 299 (275/24) & 244 (238/6) & 207 (207/0) \\ \hline
  & solver name & \textbf{\recongo} & @tigrisg & {\pariss} \\
  shortest & method & ASP & brute force/SARSA & planning \\
  & score & 238 & 232 & 213 \\ \hline
  & solver name & {\pariss} & \textbf{\recongo} & {\aiger} \\
  longest & method & planning & ASP & SAT/BMC \\
  & score & 144 & 115 & 54 \\
\end{tabular}

   \label{tbl:cmpt:result}
\end{table}

\textbf{CoRe Challenge 2022.} 
Finally, we discuss the competitiveness of our approach by empirically
contrasting it to the top-ranked solvers of the CoRe Challenge 2022~\cite{soh22:core-challenge}.
The CoRe Challenge 2022 is the first international competition on
combinatorial reconfiguration held from November 2021 to March 2022.
The competition consists of two tracks: solver track and graph track.
The solver track is divided into the following three metrics.
\begin{enumerate}
\item \emph{existent}:
  This metric is to decide the reachability of ISRP.
  Its evaluation index is the number of instances that contestants can solve.
\item \emph{shortest}:
  This metric is to find reconfiguration sequences as short as possible of ISRP.
  Its evaluation index is the number of instances that contestants can
  find the shortest sequence among all contestants.
\item \emph{longest}: 
  This metric is to find reconfiguration sequences of as long as
  possible of ISRP.
  Its evaluation index is the number of instances that contestants can
  find the longest sequence among all contestants.
\end{enumerate}
Each metric is evaluated by two indices: single-engine solvers and overall solvers.
The former index can be applied only to sequential solvers.
The latter index can be applied to all solvers, including
portfolio solvers as well as sequential solvers.
The benchmark instances are the same as ones used in our experiments above.
There are no restrictions on solvers used,
time limits, or execution environments.
Eight solvers (from seven groups) participated in the solver track of
CoRe Challenge 2022.

Table~\ref{tbl:cmpt:result} shows the results of
the top-ranked solvers of single-engine solvers track.
The columns display in order
the metric,
the solver name,
the implementation method, and
the score for each solver.
Our proposed solver {\recongo} ranked first in the shortest metric,
ranked second both in the existent and longest metrics of
single-engine solvers track.
In addition, it also ranked second in the longest metric and 
ranked third in the shortest metric of overall solvers track.
Overall,
our declarative approach can be highly competitive in performance.
On the other hand, we can observe that
many other top-ranked solvers are based on the techniques of 
bounded model checking (BMC; \cite{biere09}) and
classical planning~\cite{kautzselman92,kautzselman96}.

\textbf{Discussion.}
We discuss some more details of the results from a practical point of view.
{\recongo} showed good performance for the \code{color04} and
\code{queen} families at all metrics.
In particular, {\recongo} was able to find the shortest reconfiguration sequences
for all instances of \code{color04} at the shortest metric.
The \code{color04} family contains many instances that have
relatively short reconfiguration sequences.
In contrast, {\recongo} is less effective to the \code{power},
\code{sp}, and \code{square} families.
Those families contain many instances for which the shortest sequences
are relatively long.
The current implementation of {\recongo} relied on simple linear
search, and this issue can be improved by utilize different
search strategies such as exponential search.
{\recongo} is also less effective to the \code{grid} since
most instances of the \code{grid} are unreachable.
To resolve this issue, we will investigate the possibility of
incorporating \emph{the numeric abstraction}
used in the {\parisp} solver to our declarative approach.

 \section{Conclusion}\label{sec:conclusion}

We proposed an approach called bounded combinatorial
reconfiguration to solving combinatorial reconfiguration problems.
We presented the design and implementation of 
bounded combinatorial reconfiguration based on Answer Set Programming.
We also presented an ASP encoding of the independent set
reconfiguration problem.
We discussed the competitiveness of our approach by empirically
contrasting it to other approaches
based on the results of CoRe Challenge 2022.
The resulting {\recongo} is an ASP-based CRP solver, which is
available from
\begin{center}
\url{https://github.com/banbaralab/recongo}.  
\end{center}

Perhaps the most relevant related fields are
bounded model checking~\cite{biere09} and
classical planning~\cite{kautzselman92,kautzselman96},
in the sense of transforming a given state to another state. 
Bounded model checking in general is to study properties
(e.g., safety and liveness) of finite state transition systems and
to decide whether there is no sequence
$X_{s}=X_{0}\rightarrow X_{1}\rightarrow X_{2}\rightarrow\cdots\rightarrow X_{\ell}=X_{g}$,
for which $X_s$ is a start state and $X_g$ is an error state expressed
by rich temporal logic.
Classical planning in general is to develop action plans
for more practical applications and to decide 
whether there are sequences of actions
for which $X_s$ is a start state and $X_g$ is a goal state.
In contrast, combinatorial reconfiguration is to study
the structure and properties of solution spaces
(e.g., connectivity, reachability, and diameters)
of source combinatorial problems
and to decide whether there are reconfiguration sequences,
but $X_s$ and $X_g$ are optional.
From a broader perspective, 
combinatorial reconfiguration can involve
the task of constructing problem instances that have the maximum length of
shortest reconfiguration sequences.
Such distinctive task has been used at 
the graph track of CoRe Challenge 2022.
On the other hand,
combinatorial reconfiguration is a relatively new research field.
Therefore, the relationship between those research fields has not been
well investigated, both from theoretical and practical points of view.
We will investigate the relationship and
will explore the possibility of synergy between techniques
independently developed in those closely related research fields.

Future works include an automated transition from ASP encodings of
combinatorial programs to ones of their reconfiguration programs,
as well as an abstract interface that automatically applies
domain-independent constraints in reconfiguration problems.
Instead of ensuring there is no loop by ASP rules,
extending our algorithm to avoid loops is an important
future work for efficient longest-CRP solving.

\bibliographystyle{splncs04}
\bibliography{local,akku,lit,procs}

\end{document}